\documentclass[sigconf]{acmart}
\usepackage{subfigure}
\usepackage{booktabs}
\usepackage{multirow}
\usepackage{balance}
\usepackage[normalem]{ulem}
\useunder{\uline}{\ul}{}
\newcommand{\revise}{}

\AtBeginDocument{%
  \providecommand\BibTeX{{%
    \normalfont B\kern-0.5em{\scshape i\kern-0.25em b}\kern-0.8em\TeX}}}

\copyrightyear{2021}
\acmYear{2021}
\setcopyright{acmcopyright}
\acmConference[SIGMOD '21]{Proceedings of the 2021 International Conference on Management of Data}{June 20--25, 2021}{Virtual Event, China}
\acmBooktitle{Proceedings of the 2021 International Conference on Management of Data (SIGMOD '21), June 20--25, 2021, Virtual Event, China}
\acmPrice{15.00}
\acmDOI{10.1145/3448016.3457564}
\acmISBN{978-1-4503-8343-1/21/06}

\begin{document}
\fancyhead{}
\title{APAN: Asynchronous Propagation Attention Network for Real-time Temporal Graph Embedding}

\author{Xuhong Wang}

\orcid{1234-5678-9012}
\affiliation{%
  \institution{Shanghai Jiao Tong University}
  \city{Shanghai}
  \country{China}
  \postcode{200240}
}
\affiliation{  
\institution{Ant Group}
\city{Hangzhou}
  \country{China}
  }

\email{wang\_xuhong@sjtu.edu.cn}

\author{Ding Lyu}
\affiliation{%
  \institution{Shanghai Jiao Tong University}
  \city{Shanghai}
  \country{China}}
\email{dylan_lyu@sjtu.edu.cn}

\author{Mengjian Li}
\author{Yang Xia}
\author{Qi Yang}
\author{Xinwen Wang}
\author{Xinguang Wang}
\affiliation{%
  \institution{Ant Group}
  \city{Hangzhou}
  \country{China}
}


\author{Ping Cui}
\author{Yupu Yang}
\email{cuiping@sjtu.edu.cn}
\email{ypyang@sjtu.edu.cn}
\affiliation{%
  \institution{Shanghai Jiao Tong University}
  \city{Shanghai}
  \country{China}}

\author{Bowen Sun}
\author{Zhenyu Guo}
\authornote{Zhenyu Guo is the corresponding author to this research.}
\email{wenxi.sbw@antgroup.com}
\email{guozhen.gzy@antgroup.com}
\affiliation{%
  \institution{Ant Group}
  \city{Hangzhou}
  \country{China}
  }


\renewcommand{\shortauthors}{Xuhong, et al.}

\begin{abstract}
To capture higher-order structural features, most GNN-based algorithms learn node representations incorporating k-hop neighbors' information. Due to the high time complexity of querying k-hop neighbors, most graph algorithms cannot be deployed in a giant dense temporal network to execute millisecond-level inference. This problem dramatically limits the potential of applying graph algorithms in certain areas, especially financial fraud detection. Therefore, we propose Asynchronous Propagation Attention Network, an asynchronous continuous time dynamic graph algorithm for real-time temporal graph embedding. Traditional graph models usually execute two serial operations: first graph querying and then model inference. Different from previous graph algorithms, we decouple model inference and graph computation to alleviate the damage of the heavy graph query operation to the speed of model inference. Extensive experiments demonstrate that the proposed method can achieve competitive performance while greatly improving the inference speed. The source code is published at a Github repository.
\end{abstract}




\maketitle

\section{Introduction}
A graph is a generic mathematical language to describe complex networks, which can continue the various fields of network science applications, such as economic network, communication network,
transportation network, social network, trading network, biological network, etc. A graph $G = (V, E)$ consists of a set of nodes $V$ and edges $E$. Every single node $v \in V$ and edge $e \in E$ may have its node and edge attributes, respectively. As the application of graph data becomes more and more widespread, how to model graph data and represent nodes as low-dimensional embedding vectors for downstream tasks has become a critical concern problem of researchers. Graph neural network (GNN) has a wide range of applications and become a promising method to achieve this goal. So far, the majority of the previous graph learning works,such as DeepWalk~\cite{DBLP:conf/kdd/PerozziAS14} and SAGE~\cite{DBLP:conf/nips/HamiltonYL17}, assume that the graph is static, which means the graph is fixed and time-invariant.
\begin{figure}[t]
    \centering
    \subfigure[Continuous Time Dynamic Graph]
    {
        \label{fig:CTDG}
        \includegraphics[width=0.45\linewidth]{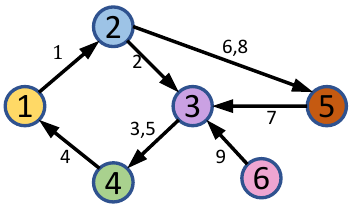}
    }
    \subfigure[Static Graph]
    {
        \label{fig:StaticGraph}
        \includegraphics[width=0.45\linewidth]{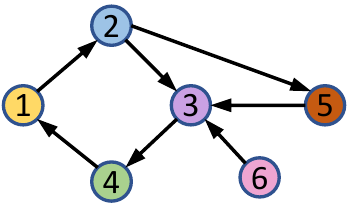}
    }
    \subfigure[Discrete Time Dynamic Graph]
    {
        \label{fig:DTDG}
        \includegraphics[width=0.9\linewidth]{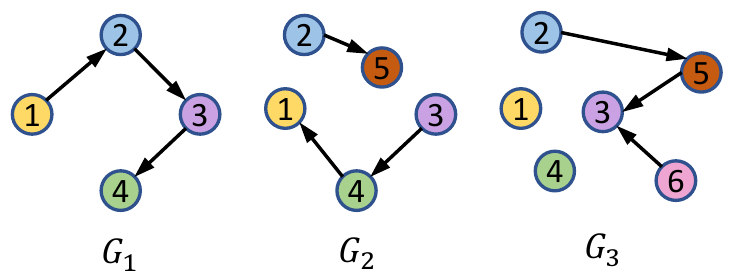}
    }
    \caption{\textbf{(a)} Continuous Time Dynamic Graph (CTDG) is the graph whose edges are labelled by timestamp. In CTDG, the interactions occur sequentially according to the order of timestamp; two nodes could have multiple interactions at different times. Whenever a node has interaction, we update the node embedding, so nodes usually have more than one dynamic embedding. \textbf{(b)} If we ignore time condition and represent CTDG as a static graph, $\textbf{v}_{4} \rightarrow \textbf{v}_{1} \rightarrow \textbf{v}_{2}$ will be considered as a valid interaction path, but it is clearly invalid with respect to time. Nodes will have only one embedding which can not represent temporal information well. \textbf{(c)} If representing CTDG as some static snapshot graphs, aka., Discrete Time Dynamic Graph (DTDG), not only some valid paths (such as $\textbf{v}_{1} \rightarrow \textbf{v}_{2} \rightarrow \textbf{v}_{5}$) will not be considered due to the snapshots partition, but also time-invalid path problem is still there.
    }
    \label{fig:Graphs}
\end{figure}

However, most real-life graph systems are dynamic: nodes and edges on graphs can appear or disappear over time, and even node attributes also may be changed. For example, in social networks, users often transfer their interest to other entities in a short time interval due to a hot spot event; in economic networks, fraudsters always commit a series of crimes suddenly and then withdraw the illicit money in the shortest time. Suppose we adopt static graph methods to model these dynamic networks, it would be simpler and more time-saving, but we can not capture evolutionary patterns of topology structures (see Figure~\ref{fig:Graphs}). Moreover, when we learn node representations on dynamic networks, we need to consider the influence of both historical events and upcoming real-time events. It is conceivable that the research difficulty on dynamic graphs is much higher than on static cases. 

\begin{figure*}[t]
    \centering
    \subfigure[Offline-deployed synchronous CTDG algorithm like TGAT~\cite{DBLP:conf/iclr/XuRKKA20}]
    {
        \label{fig:synchronous}
        \includegraphics[width=0.45\textwidth]{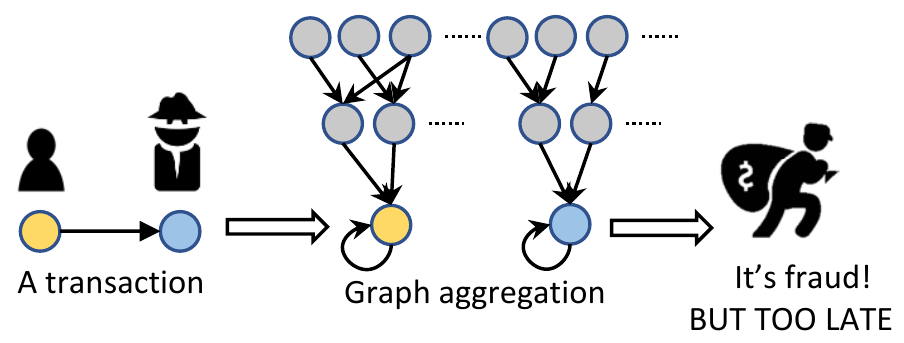}
    }
    \subfigure[Online-deployed asynchronous CTDG algorithm APAN]
    {
        \label{fig:asynchronous}
        \includegraphics[width=0.45\textwidth]{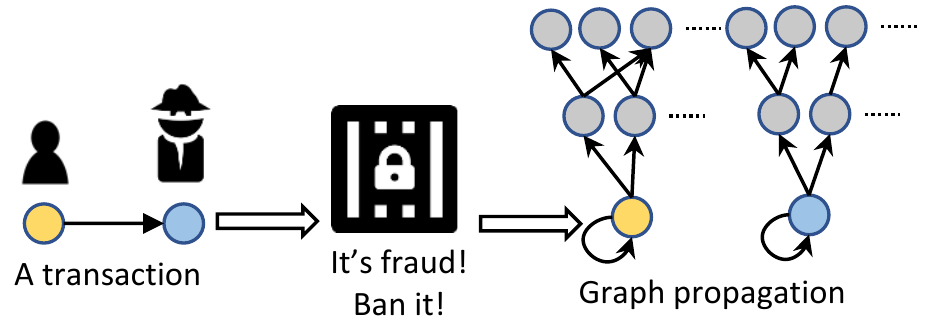}
    }
    \caption{(a) Since users cannot tolerate the high latency of neighbor query in a giant graph database, deploying a synchronous CTDG model in online payment platform is almost worthless. If we deploy it in the offline system, we might not be able to ban the accounts before fraudsters withdraw illegal fund, causing economic and reputation losses to our platform. (b) Asynchronous Propagation Attention Network (APAN) fundamentally redesigns the workflow of the CTDG algorithm. The graph querying and computation phase is transferred to the back of model inference; TGAT utilizes graph aggregation technique to model the temporal graph structure, whereas APAN uses graph propagation. APAN meets the real-time requirements of our online deployment.
    }
    \label{fig:synchronous and asynchronous}
\end{figure*}
Generally speaking, dynamic graph algorithms can be roughly divided into two categories: Discrete-Time Dynamic Graph (DTDG) and Continuous-Time Dynamic Graph (CTDG). DTDG algorithms, such as EPNE\cite{DBLP:conf/ecai/WangJSM20} and E-LSTM-D\cite{chen2019lstm}, usually transform the whole dynamic graph into a sequence of static subgraphs by time interval, a.k.a. snapshots. These methods use the discrete time window to represent the continuous-time interactions among nodes. Therefore, the performance of DTDG model is sensitive to the choice of window size and the time-variant information will be definitely lost inside the snapshot.

Recently, CTDG based graph algorithms have attracted more and more attention in the graph mining community. These paradigm of algorithms is designed to deal with batches of temporal interactions $(\textbf{v}_i,\textbf{v}_j,\textbf{e}_{ij},t)$ between nodes, which will gain more elasticity than snapshot based DTDG algorithms. Most CTDG algorithms, such as TGAT~\cite{DBLP:conf/iclr/XuRKKA20} and TGN~\cite{DBLP:journals/corr/abs-2006-10637}, model dynamic node states through temporal subgraph aggregation triggered by events.

\revise{As CTDG algorithms achieve excellent performance on learning dynamic graph embedding, they suffer from high latency during the online inference. These models usually execute two serial operations: graph querying and model inference. When a batch of interactions $\textbf{v}_i,\textbf{v}_j,\textbf{e}_{ij},t$ is coming, CTDG algorithms need to firstly visit their k-hop temporal neighbors $\mathcal{N}_{\textbf{v}_{i}}(t)$ and $\mathcal{N}_{\textbf{v}_j}(t)$. Afterwards, the model aggregate these neighbors' information to generate the node embeddings of nodes $\textbf{v}_i,\textbf{v}_j$. In real-time temporal networks, CTDG algorithms need to be deployed online and accomplish real-time inference, and it should be a millisecond-level process. 
}


\revise{Then we will give a concrete use case to show the applications of dynamic GNN methods and the problems that inspired this work. Alipay\footnote{\url{https://global.alipay.com/}} is the largest online payment App in the world, on which various transfer transactions will be carried out between users and users. At the same time, thousands of cases of embezzlement, fraud, money laundering, and gambling occur in this app every day. Such criminals often form closely connected and constantly changing graph communities (users are nodes and money transferring are edges). Therefore, graph based 
algorithms are of vital importance in these financial fraud detection task. Dynamic GNN methods provide a promising way to utilize large-scale temporal events to learn node representations, but the model inference time faces the most strict timeliness challenge. If the fraud detection system can not ban a fraudulent transaction immediately, the fraudster may escape from the platform's monitoring by withdrawing the illicit money before the system responding, causing immeasurable economic and reputation losses to the financial platform and users.
}

In addition to the rapid response to financial crimes, applying CTDG algorithms in other advantageous areas also faces various technical problems. 
\textbf{a)} When the interaction frequency suddenly increases in a short time, such as Black Friday, the graph databases will be overloaded and the entire platform can be unstable. \textbf{b)} Indeed, CTDG algorithms have the potential to capture rapid changes of nodes and graphs, but if we cannot deploy them in the online platform due to efficiency limitations, this potential cannot be maximized.


As far as we know, no effort has done to overcome these serious problems of CTDG algorithms. 
Although Sergi~et~al.~\cite{DBLP:journals/corr/abs-2010-00130} describe various GNN software and hardware acceleration schemes, they are still constrained by the basic GNN framework. 
In this work, we redesign GNN framework to decouple model inference and graph querying step so that the heavy graph query operations will not damage the speed of model inference. For the convenience of explanation, we call TGAT-like algorithms as synchronous CTDG and our APAN as asynchronous CTDG. In Figure~\ref{fig:synchronous and asynchronous}, we explain the differences between synchronous and asynchronous CTDG. \revise{Putting heavy querying and computing operations into asynchronous links can isolate the complicated algorithm from the online business decision system, and then obtain higher system stability and scalability.} Intuitively, asynchronous CTDG can meet our requirements, but designing an asynchronous CTDG is by no means as simple as adjusting the order of the graph computation phase.

Asynchronous Propagation Attention Network (APAN) is our firstly proposed model that satisfies the aforementioned asynchronous CTDG algorithm framework. APAN has two links: synchronous inference link and asynchronous propagation link. In the asynchronous link, once the interaction is completed, the detailed information of the interaction will be delivered as a "mail" to the "mailbox" of its k-hop neighbors; in the synchronous link, when the interaction occurs, APAN does not need to query the neighbors in the temporal graph, but only reads out the "mailbox" of the related nodes and generate the real-time inference instead. 

APAN has been widely tested in three real-world temporal graph datasets (including two public datasets and an industrial online payment dataset collected from Alipay. Compared with other SOTA graph deep learning models, APAN achieves significant inference speed improvements with competitive performance. 
In addition to the high inference speed and high performance, the asynchronous propagation mechanism also brings more interesting benefits to APAN. \textbf{a)} because the detailed information of interactions is stored in the mailbox, APAN has the potential of being a interpretable model. (Section~\ref{sec:Asynchronous CTDG}) \textbf{b)} APAN overcomes the common shortcoming of traditional CTDG algorithm---its sensitivity to batch size. (Section~\ref{sec:Robustness}) \textbf{c)} In some tasks, improving inference speed can also effectively improve business values. (Section~\ref{sec:Efficiency})

\section{Related Work}
\label{sec:works}
\subsection{Network Embedding}
Network embedding~\cite{cai2018comprehensive} approaches have achieved excellent performance on learning low-dimensional representations of nodes, edges, subgraphs, and the whole graph when preserving both networks' topologic properties and semantic information. There has been a burst of remarkable methods for embedding nodes of static graphs into a low-dimensional vector space. DeepWalk~\cite{DBLP:conf/kdd/PerozziAS14} firstly generates random works to sample multiplex structural information and put node sequences into a skip-gram model~\cite{DBLP:conf/nips/MikolovSCCD13} induced from Word2Vec. 
Node2Vec~\cite{grover2016node2vec} balances breadth-first sampling and depth-first sampling to obtain properties of homophily and structural equivalence. 
Moreover, LINE~\cite{DBLP:conf/www/TangQWZYM15} and SDNE~\cite{DBLP:conf/kdd/WangC016} build a similarity matrix based on 1-order and even higher-order node similarity instead of random walk strategy. Representations can also be induced from Laplacians of the adjacency matrix by non-negative matrix factorization~\cite{wang2017community,qiu2018network,qiu2019netsmf}.
However, those graph embedding method does not deal with the graph structure directly, but transforms it into a linear structure similar to text by serializing nodes and neighborhoods. Besides, random walk based network embedding method suffers from its transductive setting so that they are difficult to deal with unseen nodes.


\subsection{Graph Neural Networks}
Graph Neural Networks (GNN) is a well-known neural graph learning framework for aggregating information from neighborhoods and ego to update self-representations, which can directly process graph structure by message passing and receiving among nodes in graph. As the majority part of GNN, graph convolutional network~\cite{DBLP:conf/iclr/KipfW17} (GCN) utilizes a convolutional aggregator to collect information based on spectral methods, and SAGE~\cite{DBLP:conf/nips/HamiltonYL17} is a comprehensive improvement to expand GCN into inductive learning which also works on unknown nodes in graphs. Furthermore, graph attention networks~\cite{DBLP:conf/iclr/VelickovicCCRLB18} (GAT) introduces an attention mechanism to assign attention importance to each neighbor. Moreover, some attempts introduce GNN into encoder-decoder neural network framework. Graph AutoEncoder (GAE) and Variational GAE (VGAE)~\cite{DBLP:journals/corr/KipfW16a} use GCN to construct the encoder and decoder, so that the node embedding can be extracted in an unsupervised manner from the graph structure information as well as node attributes. Furthermore, more and more researchers try to add more skip connections into graph neural network layers to make each node aggregate more information from neighbors k hops away, which can capture long-range dependencies with Non-local Neural Networks (NLNN)~\cite{DBLP:conf/cvpr/0004GGH18}. Since the limitations of neighborhood aggregation schemes have been known, the Jump Knowledge Network~\cite{DBLP:conf/icml/XuLTSKJ18} is proposed to learn adaptive, structure-aware representations.

\subsection{Discrete-Time Dynamic Graph Learning}
There have been only a few works on representation learning on temporal networks due to complicated network evolution and dynamics. Most of them focus on learning node representations within several snapshots aggregated with edges of which timestamps are in a predefined window. This type of method is also known as the DTDG algorithm. To develop explainable models preserving evolutionary patterns, DynamicTriad\cite{zhou2018dynamic} studies on the triadic closure process to catch how open triads evolve into closed triads; SPNN\cite{meng2018subgraph} focuses on subgraphs' architecture via high-order dependencies, particularly induced subgraphs into both the input features and the model architecture; EPNE\cite{DBLP:conf/ecai/WangJSM20} divides two channels for periodic and non-periodic patterns and then incorporates both structural and temporal information. There are also several studies, such as E-LSTM-D\cite{chen2019lstm} and Know-Evolve\cite{DBLP:conf/icml/TrivediDWS17}, using an LSTM to preserve historical information and current information when learning representations. 
\begin{figure*}[t]
    \centering
    \includegraphics[width=0.9\textwidth]{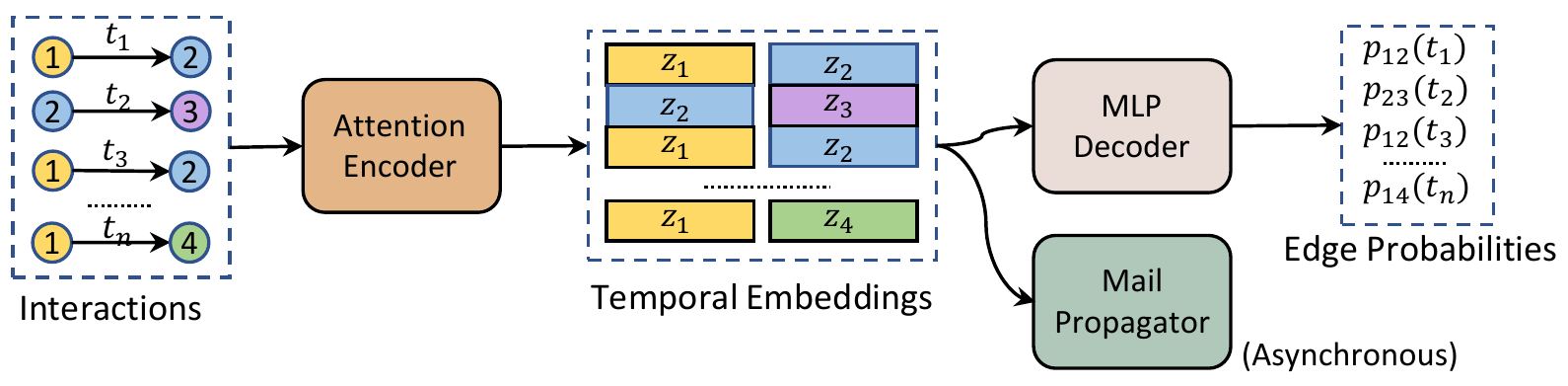}
    \caption{The overall framework of the proposed Asynchronous Propagation Attention Network (APAN). APAN can be mainly divided into three parts: Encoder, Decoder and Propagator. Note that Encoder and Decoder are in the synchronous link, and they do not need to query neighbors' information from graph database. Therefore, the time latency from the interaction occurring to the model inferring will be very short, and user will get a very smooth experience. After model inferring, the mail propagator will generate a \textit{mail} according to the interaction and then propagate it to the k-hop neighbors' \textit{mailbox} along the temporal edges.}
    \label{fig:APAN}
\end{figure*}

\subsection{Continuous-Time Dynamic Graph Learning}
Besides, on continuous-time dynamic networks, CTDNE\cite{nguyen2018continuous} suggests incorporating temporal random walks within a static skip-gram model; HTNE\cite{DBLP:conf/kdd/ZuoLLGHW18} applies a hawker process on neighborhood formation;  DynRep\cite{trivedi2019dyrep} jointly learns the topological evolution and activities between nodes in which the representation of node $v_i$ being updated after an event involving $v_i$. 

JODIE~\cite{DBLP:conf/kdd/KumarZL19} uses an RNN model to update the node memory of related nodes, whereas TigeCMN~\cite{DBLP:conf/www/ZhangBEZYL020} adopts a key-value memory network~\cite{DBLP:conf/wsdm/ChenXZT0QZ18} to update the memory. Similarly, both these two works introduce an attention architecture to read out the node memory and generate node embedding vector. However, neither JODIE nor TigeCMN learn the topology structure of the graph explicitly, because they only update the related two nodes on an edge, which means they can not visit the 2-hop neighbors directly. TGAT~\cite{DBLP:conf/iclr/XuRKKA20} leverages a time embedding kernel to a customized GAT model, gaining the capability of temporal graph encoding. Combining the advantages of JODIE and TGAT, TGN~\cite{DBLP:journals/corr/abs-2006-10637} introduces the node-wise memory into the temporal aggregate phase of TGAT. 




\section{Method}

\subsection{Definitions}
\textbf{Node embedding in static graphs}.
A static graph $\mathcal{G}=(\mathcal{V}, \mathcal{E})$, where  $\mathcal{V}=\left\{\textbf{v}_{i}, \forall i \in 1, \dots, N \right\}$ is the set of $N=|\mathcal{V}|$ nodes, $\mathcal{E} \subseteq \mathcal{V} \times \mathcal{V}$ is the set of $M=|\mathcal{E}|$ edges between nodes, where $\textbf{e}_{ij}, \forall i,j \in 1, \dots, N$ represents an edge between node $i$ and $j$. A typical Graph Neural Network (GNN) layer 
learns the representation $h_i$ of node $i$ by aggregating neighborhood information:
\begin{equation}
\begin{aligned}
\mathbf{h}_{i}^{l} &\leftarrow \sigma \cdot \mathcal{W}^{l} \cdot \left(\mathbf{h}_{i}^{l-1} || \mathbf{h}_{\mathcal{N}^{k}_{i}}^{l}\right), \\
\mathbf{h}_{\mathcal{N}^{k}_{i}}^{l} &\leftarrow \text {AGGREGATE}\left(\left\{\mathbf{h}_{j}^{l-1}, \forall j \in \mathcal{N}^{k}_{i} \right\}\right),
\end{aligned}
\end{equation}
where the parameters of $\mathcal{W}^{l}$ as well as $\text{AGGREGATE}$ are learnable, $\sigma$ is a nonlinear activation function, and $\mathcal{N}^{k}_{i}$ represents the selected k-hop neighbors of node $\textbf{v}_{i}$.

\textbf{Dynamic Graph}. 
Dynamic graphs can be classified into two categories: Discrete-time dynamic graphs (DTDG) and Continuous-time dynamic graphs (CTDG). DTDG is formed by a sequence of static graph snapshots split by time interval: $\mathcal{G}=\left\{\mathcal{G}_{1}, \mathcal{G}_{2}, \ldots, \mathcal{G}_{T}\right\}$, where $T$ is the discrete time when the snapshot is taken. CTDG is more flexible and general, which is constructed based on plenty of temporal events $\delta(t)=(\textbf{v}_i,\textbf{v}_j,\textbf{e}_{ij},t)$ order by timestamp, which means node $\textbf{v}_i$ interacts with node $\textbf{v}_j$ at time $t$. The interaction feature matrix $\textbf{e}_{ij} \in \mathcal{R}^{M \times d}$ consists of all temporal events in a CTDG, where $M$ is the number of events and $d$ is the dimension of the edge feature. Correspondingly, a CTDG can be represented as $\mathcal{G}=\left\{\delta\left(t_{1}\right), \delta\left(t_{2}\right), \ldots\right\}$. Generally, a CTDG is a multigraph, which means there might be several events at different timestamps between two nodes. 

\textbf{Node embedding in CTDG}.
Given a CTDG represented as a sequence of interaction events
$\delta(t)=(\textbf{v}_i,\textbf{v}_j,\textbf{e}_{ij},t)$, our goal is to learn a function $ f: \delta(t) \rightarrow \textbf{z}_{v_i}(t),\textbf{z}_{v_j}(t)$, where $\textbf{z}_{v_i}(t), \textbf{z}_{v_j}(t) \in \mathcal{R}^d$ respectively represent temporal embeddings of node $v_i$ and $v_j$.

\subsection{Overall Framework}

Figure~\ref{fig:APAN} provides an overview of our proposed APAN, which is the first Asynchronous CTDG algorithm for real-time temporal graph embedding. The frame of APAN is organized with an attention based encoder, an Multi-Layer Perceptron (MLP) based decoder and an asynchronous mail propagator module. 

\textbf{Synchronous Part.} When an interaction occurs, the encoder updates node embeddings according to the detail of the event, the historical embedding, and the mailbox data. Note that if a node involves in several interactions in a batch, the embedding will be generated only once. Even so, on the whole, temporal embeddings of all nodes involved in new events need to be real-time updated, which promotes the model to capture graph dynamics. Afterward, the MLP decoder will utilize these updated node embeddings to achieve downstream tasks, such as link prediction, node classification, edge classification, and node cluster. Since the encoder and the decoder are all feed-forward neural networks and we do not need to query graph neighbors in the graph database, completing these two-phase will be very fast. 

\textbf{Asynchronous Part.} After the temporal embeddings are generated, the mail propagator first creates an \textit{mail} and then propagate it to the k-hop neighbors' \textit{mailbox} along the temporal edges. The mail covers interactions associated with the sending node. Since the mail propagator is in the asynchronous link and will not harm the user experience, we can process some more complex computation in this module, such as aggregating neighbors from multiple layers or computing the subgraph statistic values.

\subsection{Attention Based Encoder}
\begin{figure}[t]
    \centering
    \includegraphics[width=0.9\linewidth]{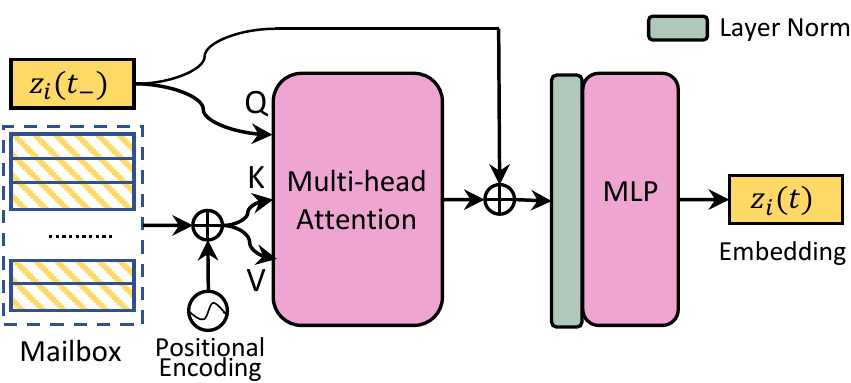}
    \caption{The encoder network of APAN is a multi-head attention module. This attention module will calculate the current node embedding $z(t) \in \mathcal{R}^d$ according to the relativity between last updated embedding $z(t-) \in \mathcal{R}^d$ and mailbox $\mathcal{M}(t) \in \mathcal{R}^{m \times d}$, where $m$ is the maximum number of mails in mailboxes. $\oplus$ means shortcut addition. }
    \label{fig:Encoder}
\end{figure}
Figure~\ref{fig:Encoder} provides the detail of the attention based encoder in APAN. This encoder introduces a classical attention architecture to create the current node embedding incorporating the last embedding $z(t-) \in \mathcal{R}^d$ and mailbox $\mathcal{M}(t) \in \mathcal{R}^{m \times d}$. $z(t-)$ indicates the node state at the last time when it participated in an interaction. Mailbox records the detailed information that the past interactions involved by neighbors, including k-hop neighbors. In this way, the encoder indirectly realizes the aggregation of temporal neighbors' information to update its node embedding. To achieve this goal, the encoder is designed by three main parts, which are positional encoding, multi-head attention, and layer normalization.

\textbf{Positional Encoding}. 
Considering the arrival order of received mails, we need to attempt positional encoding for every single mail. Because we already set the maximum number of mails in the mailbox, we can transfer the position information to the one-hot format and then feed them into the embedding look-up layer. The outputs of the embedding look-up layer are dense vectors that represent their stationary properties and are more suitable for learning by a neural network. 

For the entity mailbox $\mathcal{M}(t) = (mail_1,  mail_2, \dots, mail_m)$, the positional encoding layer combines the position information to the original mailbox matrix by
\begin{equation}
\hat{\mathcal{M}}(t)=\mathcal{M}(t) + \mathcal{P}(t)=\left[mail_{1}+p_{1}, \ldots, mail_{m}+p_{m}\right]^{\top}
\end{equation}
where $\hat{\mathcal{M}}(t), \mathcal{M}(t), \mathcal{P}(t) \in \mathcal{R}^{m \times d}$,  where $m$ is the maximum length of mailbox and $d$ is the dimension of mails. The mail dimension defaults to be the dimension of the edge feature, which we will explain in section~\ref{sec:MailProp}.

\textbf{Multi-head Attention}. 
The scaled dot-product attention~\cite{DBLP:conf/nips/VaswaniSPUJGKP17} is used as the attention module of our encoder. The hidden mechanism of an attention layer can be defined as:
\begin{equation}
\begin{aligned}
\operatorname{Attn}(\mathbf{Q}, \mathbf{K}, \mathbf{V})&=\operatorname{softmax}\left(\frac{\mathbf{Q} \mathbf{K}^{\top}}{\sqrt{d}}\right) \mathbf{V},\\
\mathbf{Q}&=z(t-)\mathbf{W}_{Q}, \\
\mathbf{K}&=\hat{\mathcal{M}}(t)\mathbf{W}_{K},
\mathbf{V}=\hat{\mathcal{M}}(t)\mathbf{W}_{V} 
\end{aligned}
\end{equation}
where $\mathbf{Q}$ denotes the `queries', $\mathbf{K}$ denotes the `keys', and $\mathbf{V}$ denotes the `values'. The dot-product attention takes a weighted sum of the entity $\mathbf{V}$ where the weights are given by the interactions of entity `$\mathbf{Q}$-$\mathbf{K}$’ pairs. The larger dot product between ’$\mathbf{Q}$-$\mathbf{K}$’ pairs reflects the greater contribution of $\mathbf{V}$ to the final output. $\mathbf{W}_{Q}, \mathbf{W}_{K}, \mathbf{W}_{V} \in \mathbb{R}^{d \times d_{h}}$ ($d_{h}$ is the dimension of the attention output) are the projection weight matrices that are employed to learn the suitable `$\mathbf{Q}$-$\mathbf{K}$-$\mathbf{V}$’ to create the expressive attention output. By using this dot-product attention, APAN model can capture the relationship between the last embedding $z(t-)$ and the mailbox $\mathcal{M}(t)$, which means the attention module can determine how to update node embedding according to the received mails from node's temporal neighbors. 

In practice, attention models always adopt multiple attention heads to form multiple subspace and force model learning different aspects of information. To build the multi-head attention module, we construct multiple attentions and concatenate them. Take four attention heads as an example:
\begin{equation}
\begin{aligned}
\text { head }_{i}&=\text { Attn }\left(Q_{i}, K_{i}, V_{i}\right), i=1, \ldots, 4 ,\\
\text {MultiHead}(Q, K, V)&=\operatorname{Concat}\left(\text {head}_{1}, \ldots, \text {head}_{4}\right) W^{O},
\end{aligned}
\end{equation}
where $W^{O} \in R^{d \times d}, head_{i} \in R^{\frac{d}{4}}$.

\textbf{Layer Normalization}. 
Since the attention outputs of different nodes are various, we need a normalization scheme to limit the mean and variance of the outputs. Layer normalization~\cite{DBLP:journals/corr/BaKH16} is the most common choice in attention models, because complex attention mechanism may disrupt the statistical distribution within a batch. If we alternatively use batch normalization, it may lead to suboptimal results. Layer normalization achieves this goal by computing the mean and variance used for normalization from all of the summed inputs to the neurons in a layer:
\begin{equation}
\begin{aligned}
a &= \text {MultiHead}(Q, K, V) + z(t-) \\
\mu&=\frac{1}{d} \sum_{i=1}^{d} a_{i},\quad
\sigma=\sqrt{\frac{1}{d} \sum_{i=1}^{d}\left(a_{i}-\mu\right)^{2}}\\
\bar{a}&=f\left(\frac{\mathrm{g}}{\sqrt{\sigma^{2}}} \odot(a-\mu)+\mathbf{b}\right)
\end{aligned}
\end{equation}
$d$ denotes the dimension of the inputs of layer normalization, $\mu$ and $\sigma$ de the mean and  variance, which are shared by all the hidden units in a layer. $\odot$ is the element-wise multiplication between two vectors. Learnable parameters $\mathrm{b}$ and $\mathrm{g}$ are defined as the bias and the gain to ensure that the normalization operation has no effect on the original information.   

After that, the outputs of layer normalization will be delivered into an MLP network to generate new temporal embeddings of nodes.

\subsection{MLP Decoder}
APAN can be widely used in a variety of downstream tasks. The attention encoder and mail propagator can be used directly without any architecture change, while the MLP decoder needs a fine-tune to adapt to different tasks. The task of the MLP decoder is to utilize the temporal node embedding to generate edge prediction for downstream tasks. For example, if we need to predict whether there will be an interaction between two nodes, the two temporal embeddings should be concatenated as $(z_i(t)||z_j(t))$ and delivered to the decoder; if we need to judge whether an edge is a fraud transaction, the two temporal embeddings and the edge feature should be concatenated as $(z_i(t)||e_{ij}(t)||z_j(t))$.

\subsection{Asynchronous Mail Propagator}
\label{sec:MailProp}
\begin{figure}[t]
    \centering
    \includegraphics[width=0.9\linewidth]{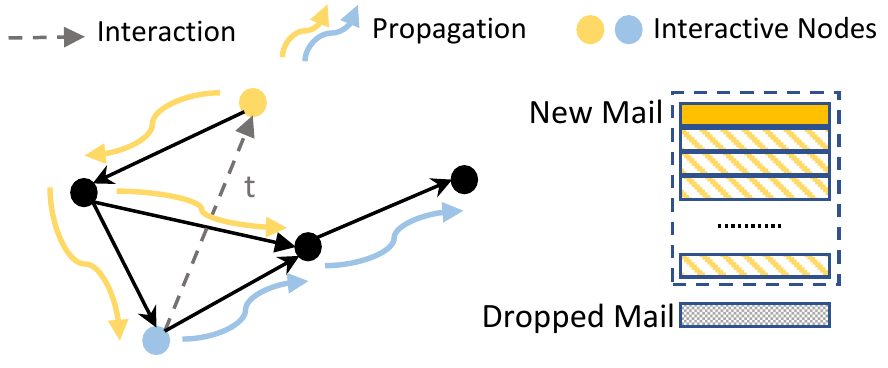}
    \caption{The workflow of the propagator of proposed APAN model. After the attention encoder generates the temporal embedding, the mail propagator first creates an interaction \textit{mail} and then propagate it to the k-hop (k=2) neighbors' \textit{mailbox} along the temporal edges. Through the mail propagation, a node can obtain the historical interaction information of its neighbors by visiting its mailbox. }
    \label{fig:Propagator}
\end{figure}
In Figure~\ref{fig:Propagator}, we demonstrate our mail propagator in the simplest way. The yellow node and the blue node have an interaction with the edge feature $e_{ij}(t)$ at time $t$, then the interaction can be characterized as a tuple $(z_i(t),e_{ij}(t),z_j(t))$, where $(z_i(t)$ and $z_j(t))$ are created by the attention encoder. The process of mail propagation in APAN can be described by two mathematical formulas below:
\begin{equation}
\begin{aligned}
Mail: mail(t)&=\phi\left(z_i(t), e_{ij}(t), z_j(t)\right), \\
Mailbox: \mathcal{M}(t)&=\psi\left(\mathcal{M}(t-), \rho\left(\left\{f\left( mail(t)\right)\right\}_{\mathcal{N}^k_{ij}}\right)\right),
\end{aligned}
\end{equation}
where $\phi(\cdot)$ is a mail-generating function to summarize the interaction details. $\mathcal{N}^k_{ij}$ represents the subgraph induced on the interactive nodes $i$ and $j$. \revise{$f(\cdot)$ is a mail-passing function that defines how a message attenuates in propagation.} $\rho(\cdot)$ is a dimension-reducing function to aggregate multiple incoming mails into one single mail. $\psi$ is an update function to update the mailbox according to the mail.


\textbf{Mail Generation ($\phi$)}.
Once a node has involved in an interaction, APAN aims at generating a mail to record what happens to this node in that interaction. A simple form of mail is just the sum of the current embeddings of the two interactive nodes and the edge feature of the current interaction, aka. $(mail(t)=z_i(t)+e_{ij}(t)+z_j(t))$. Note that the mail is also labeled by timestamp. The reason why we use summation instead of concatenation is that summation can save the memory capacity occupied by the mailbox. The disadvantage is that summation limits the dimension of node embeddings, and might obtain sub-optimal solutions in some cases.

\textbf{Temporal Neighbors Sampling ($\mathcal{N}^k_{ij}$)}.
Once a mail has been generated, we should deliver it to other nodes in order to let them know what happens to their neighbors. However, delivering one's mail to its all neighbors is inefficient. In most GNN algorithms, the node message is propagated on the sampled subgraph. The difference is that some algorithms use uniform sampling~\cite{DBLP:conf/nips/HamiltonYL17}, some algorithms use weighted sampling~\cite{DBLP:conf/kdd/YingHCEHL18}, and other algorithms use adaptive neighbor sampling~\cite{DBLP:conf/nips/Huang0RH18}. In this paper, we adapt the most-recent neighbor sampling strategy to our APAN, because CTDG methods aim at modeling the rapid change trend and updating the node embedding. Therefore, most-recent sampling is easier to restore the time-variant information, and similar experiments and conclusions can be found in that research~\cite{DBLP:journals/corr/abs-2006-10637}.

\textbf{Mail Passing ($f$)}. \revise{
After determining the propagation boundary, a.k.a, after neighbor sampling, we need a function that is used to find a reasonable attenuation or mapping mode of mails. In APAN, the mail passing function is simply set as an identification function. Note that this mail passing path is strictly obeyed the structure of interaction links between seed nodes and their neighbors. Therefore, even identification function can capture the characteristic of graph structure induced by historical interaction data.  
}

\textbf{Mail Reducing ($\rho$)}.
In practice, a node usually receives multiple mails during mail propagation. Active (high-degree) nodes usually receive more mails than inactive (low-degree) nodes. To avoid this imbalance, we use a reduction function of `mean’ operations to transform multiple mails into one. In this way, each node will receive only one mail in each single batch. This mechanism ensures that it is easy to design the next mailbox updating module. 

\textbf{Mailbox Updating ($\psi$)}.
Once a node receives a mail, its mailbox should be updated to summarize the historical state of the node's neighbors. To be as concise as possible, we adopt a first-in-first-out queue data structure to update the mailbox. Through this queue structure, the mailbox will retain the latest information and discard old mails. 

\subsection{Asynchronous CTDG framework}
\label{sec:Asynchronous CTDG}
Conceptually, the asynchronous CTDG framework we first proposed aims to solve the problem that GNN-based methods are very hard to be deployed in the millisecond-level online platform. APAN is just one of the simplest models that conform to this framework. 
Almost every module in the attention encoder and the mail propagator still have a very large room for improvement. Compared with these simple module we proposed, other more tricky modules may have more potential of improving the asynchronous CTDG framework:

\revise{
\textbf{Mailbox Mechanism}. 
We introduce the additional benefits of introducing mailboxes. Mailbox mechanism is necessary for temporal GNN models that are deployed online. In streaming systems (especially distributed streaming systems), we cannot guarantee that events will arrive in timestamp order. Therefore, it will bring instability to some machine learning models that rely on RNN dynamic updating, such as TGN and JODIE. Mailbox mechanism fixes this serious problem because the events and messages stored in the mailbox will be sorted by their timestamp when reading out.
}

\textbf{Mailbox Updating}. The key-value memory network~\cite{DBLP:conf/wsdm/ChenXZT0QZ18} framework provided a possible direction for enhancing the mailbox updating module.

\textbf{Positional Encoding}. The time embedding kernel proposed by Xu~et.~al.~\cite{DBLP:conf/iclr/XuRKKA20} can be leveraged to encode timestamps to replace the position encoding module in our APAN.

\textbf{Interpretability}. Suppose an interaction occurs at time $t$, a mail stores the detailed information which includes the node embedding $z_{i}(t)$, $z_{j}(t)$ and the edge feature $e_{ij}(t)$. Then we can use the attention weight to calculate which mail has the greatest impact on the final node embedding. This kind of interpretability is something that other models do not have because they do not store $z_{i}(t)$ and $z_{j}(t)$ but only edge features $e_{ij}(t)$.

However, a) introducing too many tricks would interfere with readers' trust in the overall architecture of our asynchronous CTDG. \revise{b) we have indeed completed the development of some related extensions, but we only tested them on the experimented datasets and did not deploy them in the actual environment.} Therefore, we will put these improvements in future work. We also look forward to the researchers from the graph machine learning community to propose better asynchronous CTDG models.

\begin{table*}[t]
\begin{tabular}{@{}lrrr@{}}
\toprule
                             & Wikipedia      & Reddit         & Alipay           \\ \midrule
Edges                        & 157474         & 672447         & 2776009          \\
Nodes                        & 9227           & 10984          & 761750           \\
Edge feature dim             & 172            & 172            & 101              \\
Nodes in train.            & 7475           & 10844          & 760289           \\
Old nodes in val. and test.    & 3131           & 10181          & 379368           \\
Unseen nodes in val. and test. & 1752           & 140            & 1461             \\
Timespan                     & 30days         & 30days         & 14days \\
Data Spilt                   & 70\%-15\%-15\% & 70\%-15\%-15\% & 10d-2d-2d       \\
Interactions with labels     & 217            & 366            & 11632            \\
Label type                   & editing ban    & posting ban    & transaction ban  \\ \bottomrule
\end{tabular}
\caption{Statistics of the datasets used in our experiments.}
\label{tab:datasets}
\end{table*}
\section{Experiments}
We test the performance of the proposed method against a variety of strong baselines (adapted for
temporal settings when possible) and competing approaches, for link prediction, node classification and edge classification tasks on two benchmarks and one large-scale industrial dataset collected from Alipay platform. The source code of our models is implemented using PyTorch and Deep Graph Library~\cite{wang2019dgl} and published at a Github repository\footnote{\url{https://github.com/WangXuhongCN/APAN}}.

\subsection{Datasets}

In this paper, we use three real-world temporal graph datasets, including two public datasets and an industrial dataset, to widely evaluate APAN's performance. Table~\ref{tab:datasets} shows the statistics of the datasets used in our experiments.

\textbf{Wikipedia}~\cite{DBLP:conf/kdd/KumarZL19} dataset is a bipartite temporal graph with \textasciitilde9,300 nodes and \textasciitilde160,000 temporal edges during one-month, where its nodes are users and wiki pages and interaction edges represent a user editing a page. Dynamic labels indicating whether a user is banned from posting. Note that there are a large number of unseen nodes (1752, 19\%) laying outside of the training dataset of Wikipedia, so Wikipedia dataset can verify the algorithms' ability in inductive learning task.

\textbf{Reddit}~\cite{DBLP:conf/kdd/KumarZL19} dataset is also a bipartite temporal graph that collects one month user interaction data, which has \textasciitilde11,000 nodes and \textasciitilde700,000 temporal edges. An interaction in Reddit dataset means a user interacting with subreddits by a post. The dynamic binary labels indicate if a user is banned from posting under a subreddit. 

\textbf{Alipay} dataset is a financial transaction dataset collected from Alipay platform, which consists of \textasciitilde760,000 nodes and \textasciitilde2,770,000 temporal edges. Each edge has a label that indicates that whether an interaction is fraud. \revise{The APAN model has been subjected to extreme and comprehensive tests for speed, memory and accuracy in the actual business platform. Due to the privacy policy, more information and the actual business indicators cannot be disclosed. Instead, we report common indicators on the Alipay dataset to characterize the business effect of the APAN algorithm. }

For Wikipedia and Reddit datasets\footnote{Wikipedia and Reddit datasets are published at \url{http://snap.stanford.edu/jodie}}, the top popular items and the most active users are considered to construct the graph; user edits consist of the textual features and are converted into 172-dimensional linguistic inquiry and word count (LIWC) feature vectors. The datasets are split with 70\%-15\%-15\% according to the interaction timestamps, where 10days-2days-2days split is used in Alipay dataset. 

Similar to paper~\cite{DBLP:conf/iclr/XuRKKA20, DBLP:journals/corr/abs-2006-10637}, since CTDG algorithms focus on the modeling the events of node and edge creation, node features are not present in any of these datasets, and we therefore assign the same zero feature vector to all nodes.

\subsection{Downstream Tasks}
In different downstream tasks, we need to choose different loss functions and metrics to evaluate the performance of various models. For example, in link prediction tasks, we use accuracy and average precision (AP) as the metrics. To train with the cross-entropy loss function, we design a time-various negative sampling strategy to construct the pairs of positive and negative samples: 
\begin{equation}
\begin{aligned}
\ell&=\sum_{\left(v_{i}, v_{j}, e_{ij}, t\right) \in \mathcal{G}}-\log \left(\sigma\left(-\mathbf{z}_{i}\left(t\right)^{\top} \mathbf{z}_{j}\left(t\right)\right)\right)\\
&-\mathbb{E}_{v_{n} \sim P_{n}(v)} \log \left(\sigma\left(\mathbf{z}_{i}\left(t\right)^{\top} \mathbf{z}_{n}\left(t\right)\right)\right),
\end{aligned}
\end{equation}
where the summation is over the training interactions between node $i$ and $j$ at time $t$, $\sigma$ is a sigmoid function and $P_{n}(v)$ is the negative sampling distribution. Note that the negative sample pool of dynamic graphs is also constantly changing dynamically. First, nodes that have never interacted cannot be sampled as negative data. Second, as the interaction continues, a historical pair of positive and negative samples may no longer valid.

In node and edge classification tasks, due to the skew of label distribution, we employ the area under the ROC curve (AUC) as the metric.

\subsection{Baselines}
As a CTDG method, the main competitors of APAN are five dynamic graph embedding methods:  CTDNE\cite{nguyen2018continuous}, DynRep\cite{trivedi2019dyrep}, JODIE~\cite{DBLP:conf/kdd/KumarZL19}, TGAT~\cite{DBLP:conf/iclr/XuRKKA20} and TGN~\cite{DBLP:journals/corr/abs-2006-10637}. In addition, we also include six static graph embedding methods to show the priority of the dynamic graph algorithms: 
DeepWalk~\cite{DBLP:conf/kdd/PerozziAS14}, Node2Vec~\cite{grover2016node2vec}, SAGE~\cite{DBLP:conf/nips/HamiltonYL17}, GAT~\cite{DBLP:conf/iclr/VelickovicCCRLB18}, GAE and VGAE~\cite{DBLP:journals/corr/KipfW16a}. We have already introduced these baselines in section~\ref{sec:works}. Note that our experiment setup closely follows TGAT~\cite{DBLP:conf/iclr/XuRKKA20} and TGN~\cite{DBLP:journals/corr/abs-2006-10637}. \revise{In Wikipedia and Reddit datasets, the results of all baselines are strictly inherited from their original papers. To be fair, we use the same data processing and splitting methods as the original paper.} For the Alipay dataset, we implement our own version according to the baseline setup described in those two papers, to explore the performance of these graph algorithms on large-scale industrial datasets.

\subsection{Configuration}
For all datasets, we use Adam optimizer with a learning rate of 0.0001,
a batch size of 200 for both training, validation and testing, a dropout rate of 0.1 and early stopping with patience of 5. The number of attention heads is set as 2 and the message passing layer is 2. For the MLP net in the encoder and decoder, we employ two-layer feedforward neural network with a hidden size of 80. Note that these parameters mentioned above are all taken from original papers of TGAT and TGN, and we did not employ complex hyper-parameter tuning to improve APAN results.

The node embedding dimension of APAN is fixed as the original edge feature dimension, so it is not a hyper-parameter. The numbers of mailbox slots and sampled neighbors are all set as 10 for all three datasets. In subsequent experiments, we will prove that APAN method is not sensitive to hyper-parameters. As long as the parameters are set within a reasonable range, APAN will hardly result in catastrophic performance.
\subsection{Results}
\begin{table}[t]
\begin{tabular}{@{}ccccc@{}}
\toprule
\multirow{2}{*}{} & \multicolumn{2}{c}{Wikipedia}                           & \multicolumn{2}{c}{Reddit}                              \\ \cmidrule(l){2-5} 
          & Accuracy    & AP          & Accuracy    & AP          \\ \midrule
GAE       & 72.85 (0.7) & 91.44 (0.1) & 74.31 (0.5) & 93.23 (0.3) \\
VAGE      & 78.01 (0.3) & 91.34 (0.3) & 74.19 (0.4) & 92.92 (0.2) \\
DeepWalk  & 76.67 (0.5) & 90.71 (0.6) & 71.43 (0.6) & 83.10 (0.5) \\
Node2vec  & 78.09 (0.4) & 91.48 (0.3) & 72.53 (0.4) & 84.58 (0.5) \\
GAT       & 87.34 (0.3) & 94.73 (0.2) & 92.14 (0.2) & 97.33 (0.2) \\
SAGE & 85.93 (0.3) & 93.56 (0.3) & 92.31 (0.2) & 97.65 (0.2) \\
CTDNE     & 79.42 (0.4) & 92.17 (0.5) & 73.76 (0.5) & 91.41 (0.3) \\
DyRep     & 87.77 (0.2) & 94.59 (0.2) & 92.11 (0.2) & 97.98 (0.1) \\
JODIE     & 87.04 (0.4) & 94.62 (0.5) & 90.91 (0.3) & 97.11 (0.3) \\
TGAT              & 88.14 (0,2)                & 95.34 (0.1)                & {\ul \textit{92.92 (0.3)}} & 98.12 (0.2)                \\
TGN               & {\ul \textit{89.51 (0.4)}} & \textbf{98.46 (0.1)}       & 92.56 (0.2)                & {\ul \textit{98.70 (0.1)}} \\
APAN              & \textbf{90.74 (0.1)}       & {\ul \textit{98.12 (0.2)}} & \textbf{94.34 (0.1)}       & \textbf{99.22 (0.2)}       \\ \bottomrule
\end{tabular}
\caption{In link prediction task, we show the average APs and accuracies in \% with StdDevs (over 10 random seeds). Note that the best results are typeset in \textbf{bold} and the second bests are highlighted with {\ul underline}. }
\label{tab:linkpredresults}
\end{table}

\begin{table}[t]
\begin{tabular}{@{}cccc@{}}
\toprule
\textbf{}     & \multicolumn{2}{c}{Node classification}                 & Edge classification           \\ \cmidrule(l){2-4} 
\textbf{}     & Wikipedia                  & Reddit                     & Alipay                        \\ \midrule
GAE           & 74.85 (0.6)                & 58.39 (0.5)                & \textbackslash{}              \\
VGAE          & 73.67 (0.8)                & 57.98 (0.6)                & \textbackslash{}              \\
GAT           & 82.34 (0.8)                & 64.52 (0.5)                & 69.47 (0.4)                   \\
SAGE     & 82.42 (0.7)                & 61.24 (0.6)                & 67.91 (0.5)                   \\
CTDNE         & 75.89 (0.5)                & 59.43 (0.6)                & \textbackslash{}              \\
DyRep         & 84.59 (2.2)                & 62.91 (2.4)                & 65.09 (1.0)                   \\
JODIE         & 83.17 (0.5)                & 59.90 (2.1)                & 81.89 (0.7)                   \\
\textit{TGAT} & \textit{83.69 (0.7)}       & {\ul \textit{65.56 (0.7)}}                & 77.84 (0.9)                   \\
TGN           & {\ul \textit{88.56 (0.3)}} & \textbf{68.63 (0.7)}       & \textit{\textbf{84.01 (0.9)}} \\
APAN          & \textbf{89.86 (0.3)}       & 65.34 (0.4) & {\ul \textit{83.37 (0.7)}}    \\ \bottomrule
\end{tabular}
\caption{In dynamic edge/node classification task, we show the average AUCs in \% with StdDevs (over 10 random seeds). }
\label{tab:fraudresults}
\end{table}

Table~\ref{tab:linkpredresults} shows the link prediction experiment results of our APAN and eleven SOTA baselines. It is obvious that almost all methods based on dynamic graphs outperform the static graph methods. Unsupervised graph embedding approaches, such as GAE, DeepWalk, Node2Vec and CTDNE, have a bad performance, because embedding learned by those methods is task agnostic, which has a limited and indirect contribution to downstream tasks. Our APAN achieves competitive performance comparing to other SOTA methods. Especially in the Reddit dataset, APAN demonstrates an amazing performance than other methods. 

The asynchronous CTDG algorithms like APAN have their own advantages that the traditional algorithm does not have. Synchronous CTDG methods usually use an update function to create the node once the node interacts, whereas, in asynchronous CTDG, the nodes‘ mailboxes are updated as long as their neighbors participate in an interaction. In other words, the node update frequency in the asynchronous CTDG algorithm is higher than that in the synchronous CTDG. It is this difference that makes APAN have the more powerful capability in dynamic graph embedding. We can also draw similar conclusions from the results of node or edge classification tasks in Table~\ref{tab:fraudresults}.

Note that the structure and hyperparameters used in this study proved to be sufficient for our applications, although they can still be improved. The main improvement of APAN is that it greatly improves the inference speed, and APAN's algorithm architecture is particularly suitable for online deployment on the Internet platform. 

\subsection{Efficiency}
\label{sec:Efficiency}
\begin{figure}[t]
    \centering
    \includegraphics[width=0.9\linewidth]{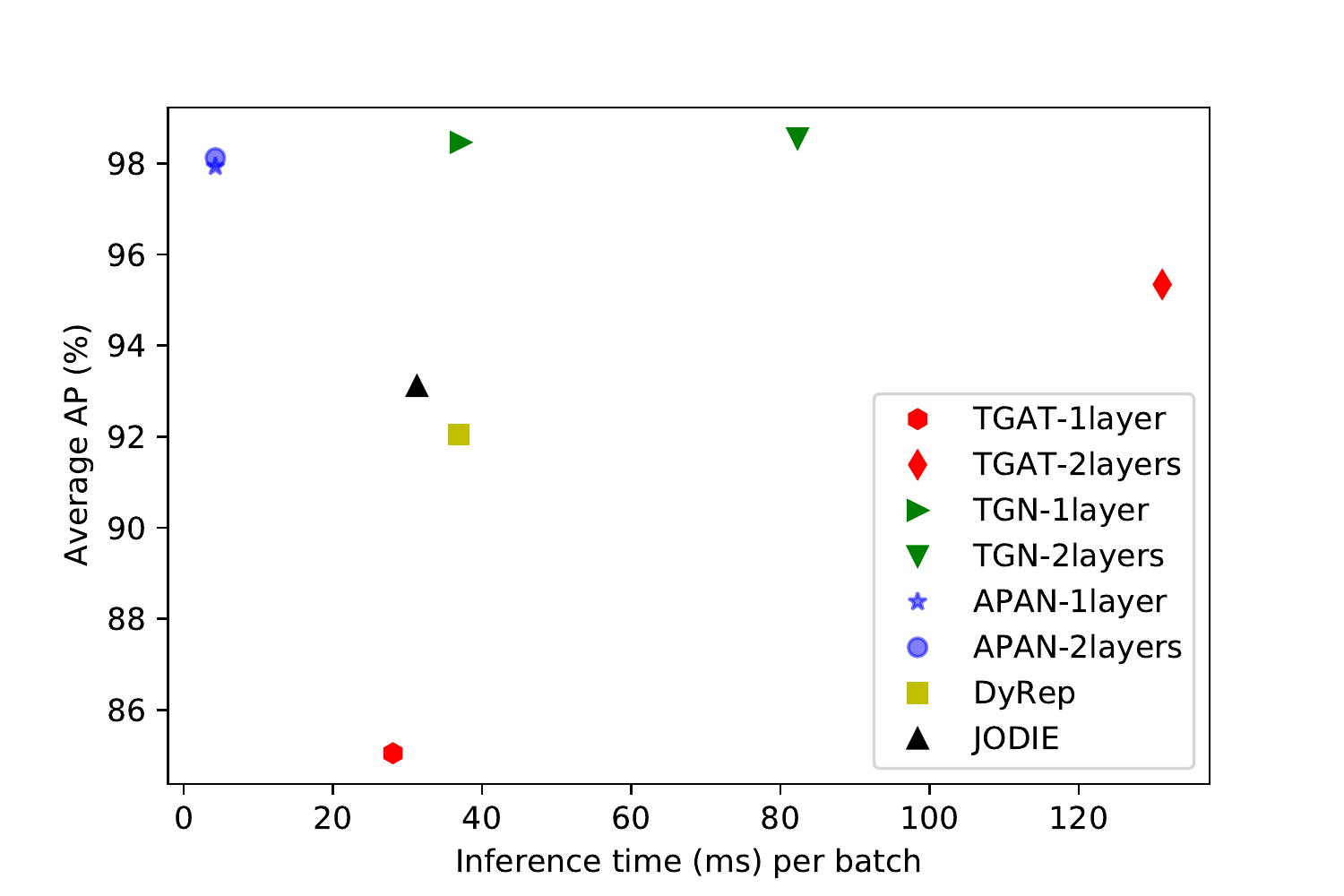}
    \caption{The AP (\%) metric and inference speed (ms) per batch in Wikipedia dataset, in link prediction task. Each batch has 200 interactions. The closer to the upper left corner, the better the performance of the model. APAN is 8.7$\times$ faster than TGN and has almost the same testing result.}
    \label{fig:inferspeed}
\end{figure}

\begin{figure}[t]
    \centering
    \includegraphics[width=0.9\linewidth]{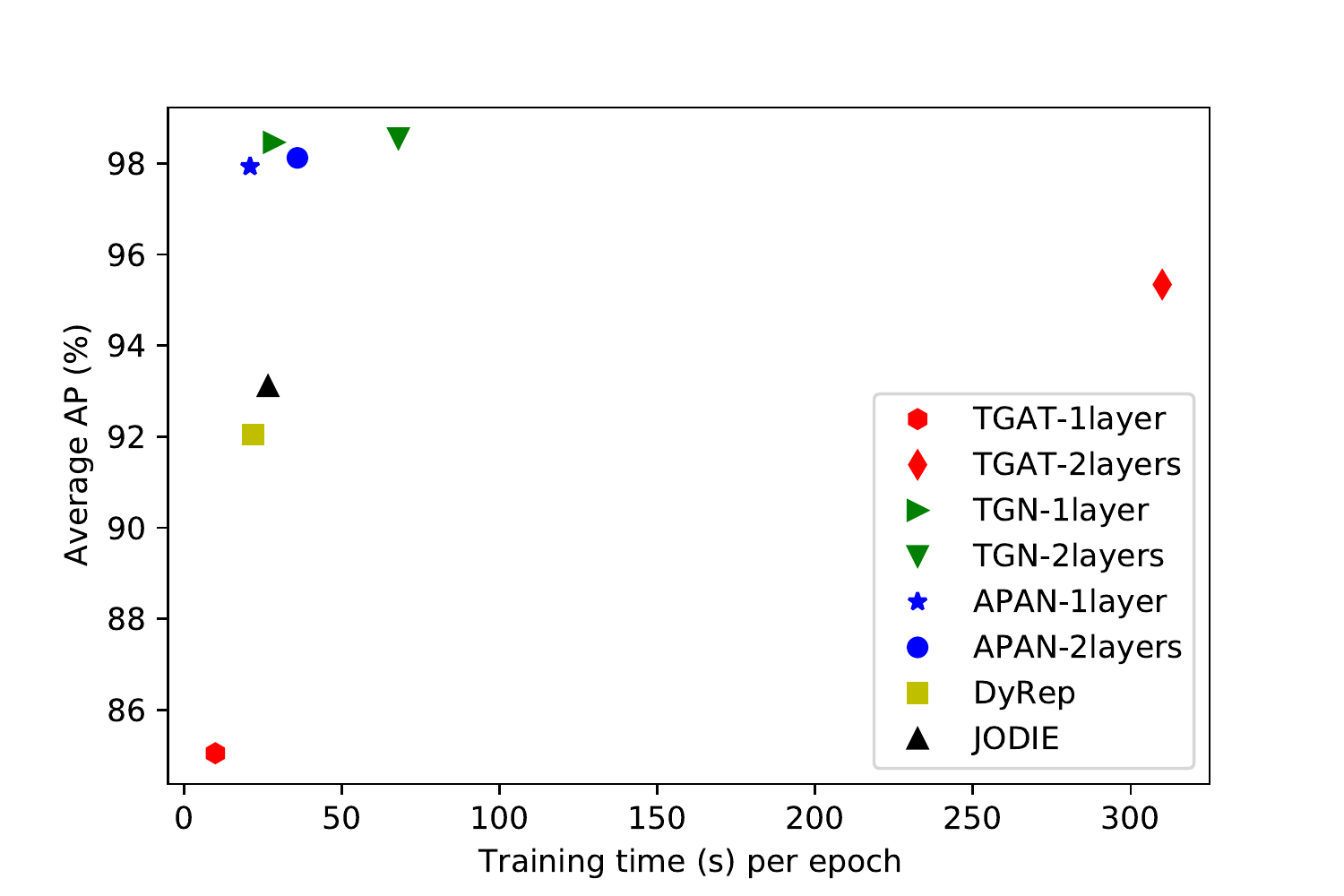}
    \caption{\revise{The AP (\%) metric and training speed (s) per epoch in Wikipedia dataset, in link prediction task. Each batch has 200 interactions. The closer to the upper left corner, the better the performance of the model. In the training phase, APAN has almost the same testing result and speed as TGN. }}
    \label{fig:trainspeed}
\end{figure}

In this section, we compare the Efficiency of APAN with other baselines. The speed of APAN in the inference and training phase are shown in Figure~\ref{fig:inferspeed} and Figure~\ref{fig:trainspeed}, respectively. The experiments run on a Linux PC with an Intel Core i7-7820X CPU (8 cores, 3.60GHz) and a 12 GB NVIDIA TITAN X (Pascal) GPU. The implement version of these models is coming from two public repositories\footnote{\url{https://github.com/twitter-research/tgn}}\footnote{\url{https://github.com/StatsDLMathsRecomSys/Inductive-representation-learning-on-temporal-graphs}}. 

In the real-world online Internet platform, the online inference time of a model is more important than training time. Take the Alipay anti-fraud system as an example, a transaction can only be executed if it passes the anti-fraud system. Long inference time will take up too much computing resource and cause the instability of the online inference engine, but APbAN overcomes this problem by putting most of the calculations on the asynchronous link. Moreover, if the inference time is too long, it will greatly damage the user experience. Therefore, a model with low inference delay will greatly increase the return on investment and increase the model's business value.

We conducted runtime experiments to simulate the average waiting time required for a batch of interactions to pass CTDG algorithms. Note that we only calculate the time from the interaction occurring to the model inference, not including the time on APAN's asynchronous link. 

In the inference phase, APAN is 8.7$\times$ faster than TGN and has almost the same testing result. JODIE and DyRep are limited by the expressive ability, so their performance lags behind APAN. With the increase in the number of layers, the performance of TGN and TGAT is improved, but the inference speed is also greatly reduced; APAN's inference speed will not change with the layers, because of its asynchronous mail propagation mechanism. It means that we can apply more complex network computation in APAN to further enhance the performance under the limit of real-time inference. 

\revise{Note that in a real system, the speed increase of APAN is much greater than 8.7 times. Because in our implementation, the whole graph is stored in the single PC memory, whereas it is stored in a distributed graph database in a real platform. The reading efficiency of large-scale distributed databases will be the bottleneck of the entire system, while it is not when using the single-machine memory. }

\revise{In the training phase, APAN has almost the same testing result and speed as the current fastest algorithm TGN. The reason is that, in the training phase, APAN is very similar to other CTDG algorithms. APAN just exchanged the calculation order without introducing additional calculations. Besides, in our implementation, the underlying efficient information dissemination mechanism is adopted, so increasing the number of layers will not result in a significant increase in training time. }

\subsection{Robustness}
\label{sec:Robustness}

\textbf{Batch size}.

\begin{figure}[t]
    \centering
    \includegraphics[width=0.9\linewidth]{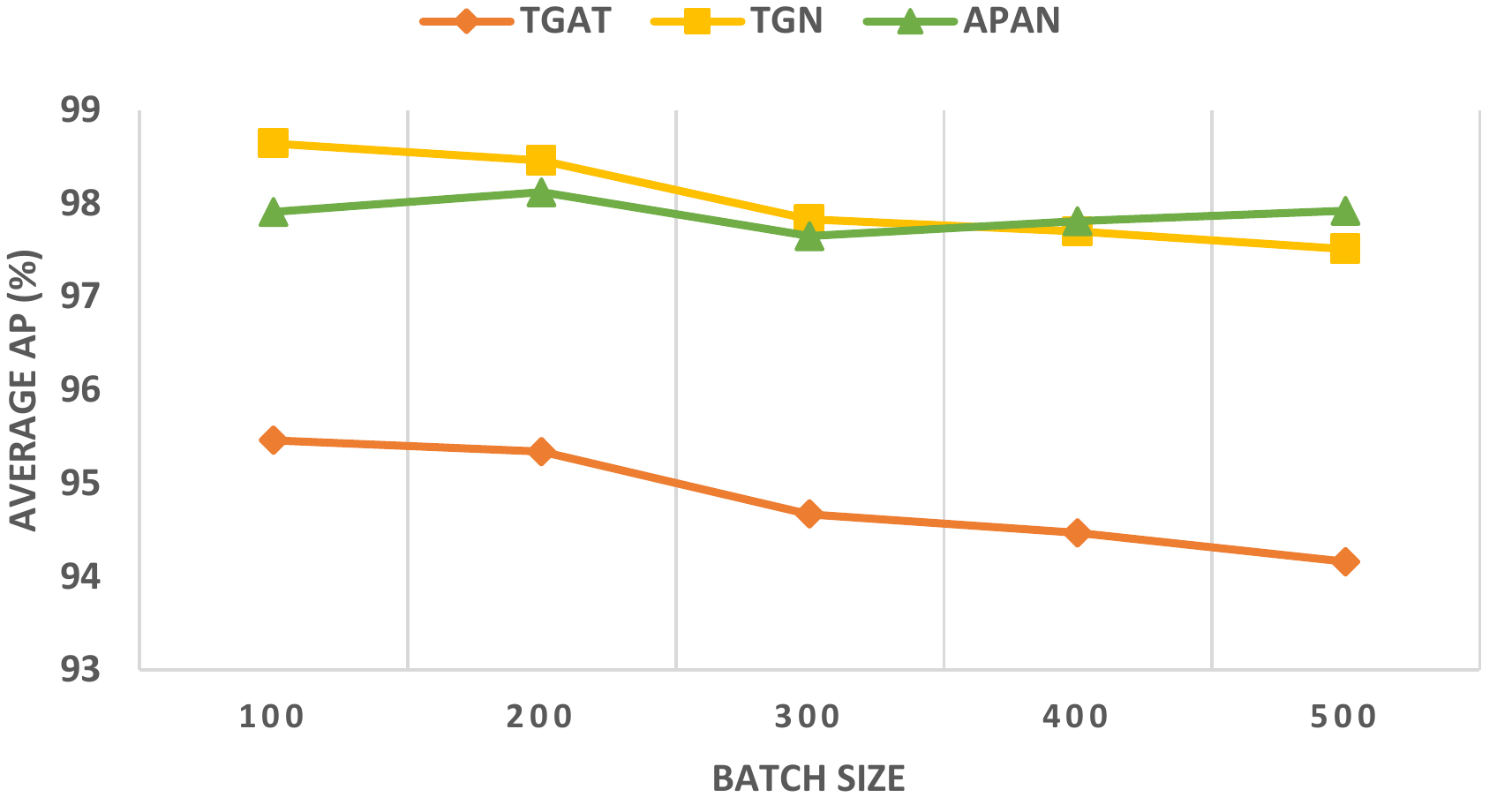}
    \caption{The relationship between batch size and the performance in Wikipedia dataset, in link prediction task. In the real world, platforms may need to handle thousands of events each batch. Thanks to the asynchronous propagation, APAN is not sensitive to the training batch size. However, the performances of TGAT and TGN decrease as the batch size increases.}
    \label{fig:batchsize}
\end{figure}
In addition to the high inference speed, the asynchronous propagation mechanism also brings more interesting benefits to APAN: robustness to batch size. The common drawback of the CTDG algorithms is their sensitivity to batch size. From Figure~\ref{fig:batchsize}, we conclude that the larger the batch size, the worse the algorithm performance. 

The most desirable condition of the CTDG algorithms is to update the node triggered by a single event, that is, the batch size is equal to 1. Suppose a batch starts at $t_0$ and interaction happens at time $t$, the most latest interactions between $t_0$ and $t$ are lost because CTDG models assume that the events in a batch arrive simultaneously. For this reason, the larger the batch size, the more information is lost, and the worse the performance of CTDG models.

To ensure good performance, TGAT and TGN adopted a batch size as small as 200 in the 700-thousand-level Reddit dataset. However, in real-world cases, Internet platforms may need to judge thousands of events each batch to adapt to business needs. We need a CTDG algorithm that is not sensitive to batch size.

Fortunately, asynchronous CTDG does not need to view the latest interactions. Synchronous CTDG models query the latest temporal subgraph when the interaction happens, whereas APAN first outputs the embedding and then queries the subgraph. \revise{As a result, APAN loses the latest interactive information of nodes. Let's use time series forecasting to give a vivid example. Given a time series $x(1), \dots,x(t-2), x(t-1), x(t)$, synchronous CTDG models capture the relationship $x(t-1) \rightarrow x(t)$, while APAN maps $x(t-2) \rightarrow x(t)$. As long as the change trend of the node is continuous, it is feasible to use $x(t-2)$ to predict $x(t)$.} 

Therefore, APAN is forced to learn how to create inference without using the latest subgraph.  
Asynchronous propagation mechanism effectively avoids the impact of batch size on performance. \revise{Besides, APAN is more tolerant of system delay, because the latest information will not arrive on time in this case. }


\begin{figure}[t]
    \centering
    \includegraphics[width=0.8\linewidth]{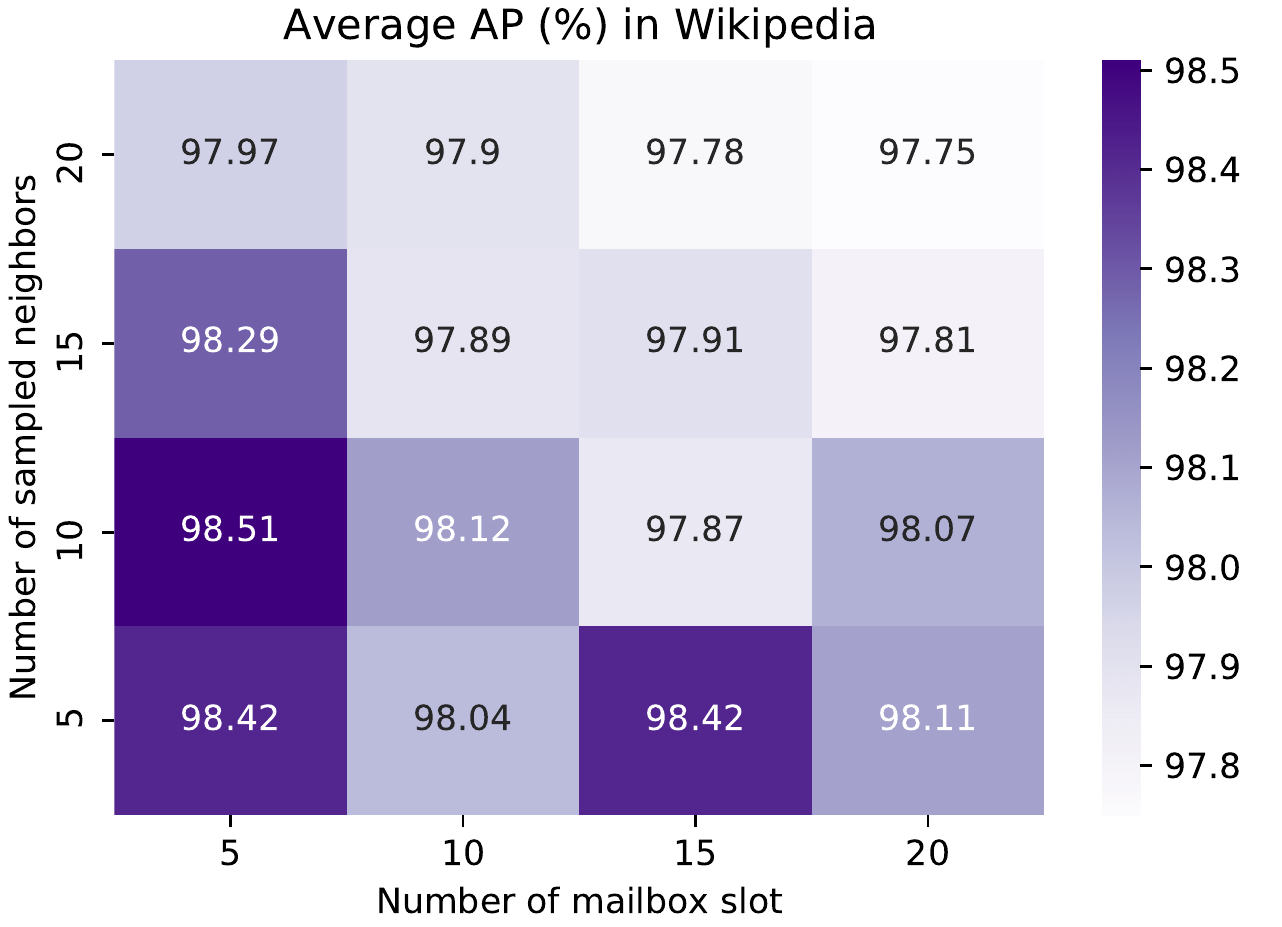}
    \caption{The average AP (\%) metrics (over 10 runs), in link prediction task, with different numbers of sampled neighbors and mailbox slots in Wikipedia dataset. In the grid search experiment on these two parameters, APAN shows its strong robustness. Its best and worst results fluctuate only 0.6\%. }
    \label{fig:mail-degree}
\end{figure}

\textbf{Number of sampled neighbors and mailbox slots.}

In Figure~\ref{fig:mail-degree}, we show the APAN's robustness to the two important hyper-parameters: number of sampled neighbors and mailbox slots. In these 16 results, the best and worst results fluctuate only 0.6\%. Although APAN is not sensitive to these two parameters, we still analyze the reasons for the performance difference. \textbf{a)} Number of mailbox slots. We can conclude that a small number of slots is enough, because the CTDG model only needs to refer to the short-term interactions to make inference in link prediction task (a similar conclusion is presented in 
TGAT~\cite{DBLP:conf/iclr/XuRKKA20}). A mailbox with large slots number will store redundant information and make the model difficult to learn.
\textbf{b)} Number of sampled neighbors is the most common parameter in GNN model. If too many neighbors are aggregated, the model may not be able to distinguish the representation after aggregation; if too few neighbors are sampled, important neighbors may be missed.

\revise{
Besides, we also explain the concerns that readers may have about the extra memory usage of the mailbox. 
a) Experiments in Figure \ref{fig:mail-degree} have proved that a very small number of mailbox slots can achieve competitive results. b) The memory usage is only related to the number of nodes, which is limited in most scenarios, whereas the number of interactive edges is unlimited (e.g., Reddit dataset has 10k nodes and 680k edges). Since the GNN models need to store a large amount of historical interactive edge information, the mailbox mechanism is not a memory bottleneck of the entire system.
}

In summary, APAN is a very robust model because it is not sensitive to the major hyperparameters, people can spend less energy to deploy APAN model on their own tasks.

\section{Conclusion}
In this paper, we proposed Asynchronous Propagation Attention Network (APAN), an asynchronous CTDG algorithm framework for real-time temporal graph embedding. APAN aims at transforming traditional CTDG algorithms to adapt to online deployment and real-time inference on Internet platforms. Our results from extensive experiments demonstrate that the proposed APAN can achieve competitive performance with 8$\times$ inference speed improvement. In the future, we will explore more opportunities for the proposed asynchronous CTDG framework.

\section{Broader Impact}
\revise{
This paper presents a new dynamic graph neural networks approach for super fast inference. As far as we know, APAN is the first GNN algorithm that can achieve millisecond-level and could help achieve super large-scale inference within the online distributed graph database. It may enhance the industry’s future design solutions of how to adapt GNN models in recommender systems, financial systems, social networks and so on. 
}
\begin{acks}
This work was supported by Ant Group through Ant Research Intern Program. This research is also supported by National Natural Science Foundation of China (No. 51777122).

\revise{In order to comply with the company's data privacy policy, the industrial dataset we used in this article follows the following statement.  \newline
1) The Alipay dataset does not contain any Personal Identifiable Information (PII). \newline
2) The Alipay dataset is desensitized and encrypted.\newline
3) Adequate data protection was carried out during the experiment to prevent the risk of data copy leakage, and the dataset was destroyed after the experiment.\newline
4) The Alipay dataset is only used for academic research, it does not represent any real business situation.
}
\end{acks}

\bibliographystyle{ACM-Reference-Format}
\balance
\bibliography{APAN}


\end{document}